\documentclass[journal]{IEEEtran}
\usepackage{times}
\usepackage{epsfig}
\usepackage{graphicx}
\usepackage{amsmath,amsfonts}
\usepackage{amssymb}
\usepackage{array}
\usepackage{multirow}
\usepackage{subcaption}
\usepackage{color}
\usepackage{float}
\usepackage{graphicx}
\usepackage{mathtools}
\usepackage{booktabs}
\usepackage{hyperref}
\usepackage{mathtools}
\usepackage{dblfloatfix}
\usepackage{cite}
\usepackage{graphics}

\newcases{Rcases}{$\mskip\thickmuskip$}{%
  \hfil$\m@th{##}$}{$\m@th{##}$\hfil}{.}{\rbrace}

\usepackage[font=scriptsize]{caption}  
\usepackage[normalem]{ulem}
\usepackage{xcolor}
\definecolor{darkgreen}{rgb}{0,.4,0}
\definecolor{darkcyan}{rgb}{0,.4,.4}
\newcommand{\REMOVE}[1]%
          {{\color{blue}\sout{#1}}}

\newcommand{\COMMENT}[1]%
          {{\color{darkgreen}\textbf{{Editor: }} {#1}}}

\definecolor{forest!}{RGB}{29,110,34}

\definecolor{industrial!}{RGB}{239,100,154}
\definecolor{lowplants!}{RGB}{18,222,49}
\definecolor{allotment!}{RGB}{110,225,175}
\definecolor{residential!}{RGB}{196,202,194}
\definecolor{commercial!}{RGB}{220,207,14}
\definecolor{soil!}{RGB}{185,182,158}
\definecolor{water!}{RGB}{13,123,206}
\definecolor{healthygrass!}{RGB}{124,222,19}
\definecolor{stressedgrass!}{RGB}{220,214,15}
\definecolor{syntheticgrass!}{RGB}{110,222,225}
\definecolor{tree!}{RGB}{102,185,124}
\definecolor{road!}{RGB}{62,57,20}
\definecolor{highway!}{RGB}{0,0,0}
\definecolor{railway!}{RGB}{102,34,114}
\definecolor{parkingL1!}{RGB}{69,202,176}
\definecolor{parkingL2!}{RGB}{100,239,156}
\definecolor{tenniscourt!}{RGB}{69,255,69}
\definecolor{runningtrack!}{RGB}{151,115,205}

\begin{document}

\title{Spatial Gated Multi-Layer Perceptron for Land Use and Land Cover Mapping}

\author{
 Ali Jamali,\textsuperscript{*}
 Swalpa Kumar Roy,\textsuperscript{*}~\IEEEmembership{Student Member,~IEEE},
 Danfeng Hong,~\IEEEmembership{Senior Member,~IEEE}, \\
 Peter M Atkinson,
 and
 Pedram Ghamisi,~\IEEEmembership{Senior Member,~IEEE}
 
\thanks{This research was funded by the Institute of Advanced Research in Artificial Intelligence (IARAI). (\textit{Corresponding author: Pedram Ghamisi})}
\thanks{A. Jamali is with the Department of Geography, Simon Fraser University, British Columbia 8888, Canada (e-mail: alij@sfu.ca).}
\thanks{S. K. Roy is with the Department of Computer Science and Engineering, Alipurduar Government Engineering and Management College, West Bengal 736206, India (e-mail: swalpa@cse.jgec.ac.in).}
\thanks{D. Hong is with the Aerospace Information Research Institute, Chinese Academy of Sciences, 100094 Beijing, China. (e-mail: hongdf@aircas.ac.cn).}
\thanks{P. M. Atkinson is with the Faculty of Science and Technology, Lancaster University, Lancaster, U.K. (e-mail:pma@lancaster.ac.uk ).}
\thanks{P. Ghamisi is with the Helmholtz-Zentrum Dresden-Rossendorf (HZDR), Helmholtz Institute Freiberg for Resource Technology, 09599 Freiberg, Germany, and is also with the Institute of Advanced Research in Artificial Intelligence (IARAI), 1030 Vienna, Austria (e-mail: p.ghamisi@gmail.com).}
\thanks{(* indicates these two authors contributed equally to the work.)}
}

\markboth{Submitted to IEEE Journal}
 {Jamali \MakeLowercase{\textit{et al.}}: Bare Demo of IEEEtran.cls for Journals}

\maketitle

\begin{abstract}

Convolutional Neural Networks (CNNs) are models that are utilized extensively for the hierarchical extraction of features. Vision transformers (ViTs), through the use of a self-attention mechanism, have recently achieved superior modeling of global contextual information compared to CNNs. However, to realize their image classification strength, ViTs require substantial training datasets. Where the available training data are limited, current advanced multi-layer perceptrons (MLPs) can provide viable alternatives to both deep CNNs and ViTs. In this paper, we developed the SGU-MLP, a learning algorithm that effectively uses both MLPs and spatial gating units (SGUs) for precise land use land cover (LULC) mapping. Results illustrated the superiority of the developed SGU-MLP classification algorithm over several CNN and CNN-ViT-based models, including HybridSN, ResNet, iFormer, EfficientFormer and CoAtNet. The proposed SGU-MLP algorithm was tested through three experiments in Houston, USA, Berlin, Germany and Augsburg, Germany. The SGU-MLP classification model was found to consistently outperform the benchmark CNN and CNN-ViT-based algorithms. For example, for the Houston experiment, SGU-MLP significantly outperformed HybridSN, CoAtNet, Efficientformer, iFormer and ResNet by approximately 15\%, 19\%, 20\%, 21\%, and 25\%, respectively, in terms of average accuracy. The code will be made publicly available at \url{https://github.com/aj1365/SGUMLP}

\end{abstract}

\begin{IEEEkeywords}
Attention mechanism, image classification, spatial gating unit (SGU), vision transformers.
\end{IEEEkeywords}

\IEEEpeerreviewmaketitle

\section{Introduction}
\label{sec:intro}

\IEEEPARstart{L}and use and land cover (LULC) is one of the most significant indicators of anthropogenic interaction with the natural environment. Massive growth in LU because of forest destruction, urbanization and soil erosion has altered the global landscape and caused greater stress on natural ecosystems across the world \cite{Yang680}. Modelers' dominant perception of cities is of a sophisticated structure with characteristics like occurrence, self-organization and non-linear relationships. Urbanization and growing urban populations have led to significant scientific debate since they lead to substantial shifts in agricultural use and environmental degradation \cite{HAASE201292}. Analysis of urban growth, including intense growth in urban sprawl, is essential for understanding its environmental consequences, as well as promoting the adoption of more sustainable forms of urban expansion. As a result, it is essential to organize and structure the process of LULC change in natural ecosystems.

Precise LULC mapping serves as the foundation for non-monetary assessment and is generally obtained by integrating a machine learning or deep learning algorithm with imagery from remote sensing. Deep learning models are capable of extracting adaptively the most important features from data using a data-driven approach. During the training phase, these models can achieve an effective parametric configuration by simultaneously training the associated classification model. This greatly enhances their ability to accurately represent complex data and avoid ambiguity \cite{FAN20249}. Deep learning models have been progressively used for LULC mapping in recent years \cite{Ghamisi2156, ahmad2021hyperspectral}. In particular, Convolutional Neural Networks (CNNs) are widely utilized models for hierarchical feature extraction. Due to their self-attention system, vision transformers (ViTs) can model global contextual information more effectively than CNNs \cite{roy2022multimodal}, but they require larger training datasets to maximize image classification accuracy. On the other hand, where fewer training data are available, current advanced Multi-layer Perceptrons (MLPs) can be used as an alternative to both deep CNNs and ViTs \cite{Tolsti2021}.

In this paper, we develop and propose an SGU-MLP, a deep learning classifier that employs MLPs and a spatial gating unit (SGU) for accurate LULC modeling. The SGU concept enables the algorithm to efficiently characterize complex spatial interactions across input data tokens without the use of positional information embedding as utilized in popular ViTs. The SGU-MLP model's final layer employs a structure entirely composed of multi-layer perceptrons (MLPs), eliminating the requirement for CNNs or ViTs and, consequently, minimizing the necessity for extensive training data.

This Letter introduces the SGU-MLP in Section~\ref{sec:prop}, illustrates the experiments and analyses the results in Section~\ref{sec:exp}, and highlights the concluding remarks in Section~\ref{sec:con}.

\begin{figure*}[!ht]
\centering
\includegraphics[clip=true, trim= 20 250 05 10, width=0.63\textwidth]{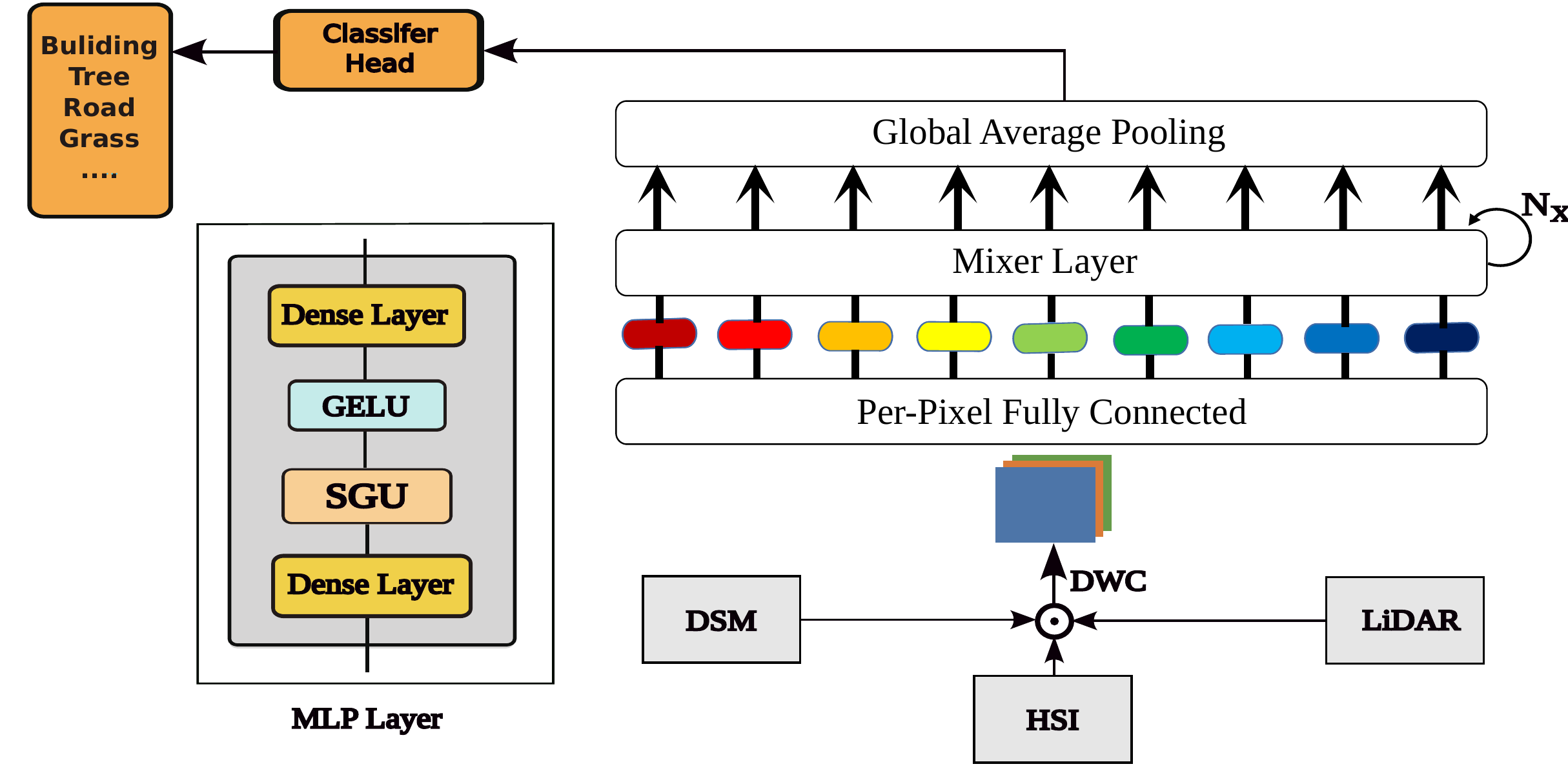}
\caption{Graphical representation of spatial gated multi-layer perceptron framework for land use and land cover classification. The MLP-Mixer layer includes two MLPs to extract spatial information. $\odot$ represents channel-wise concatenation.}
\label{fig:prop}
\end{figure*}

\section{Proposed Classification Framework}
\label{sec:prop}

\begin{figure*}[!ht]
\centering
\includegraphics[clip=true, trim= 20 600 20 10, width=0.7\linewidth]{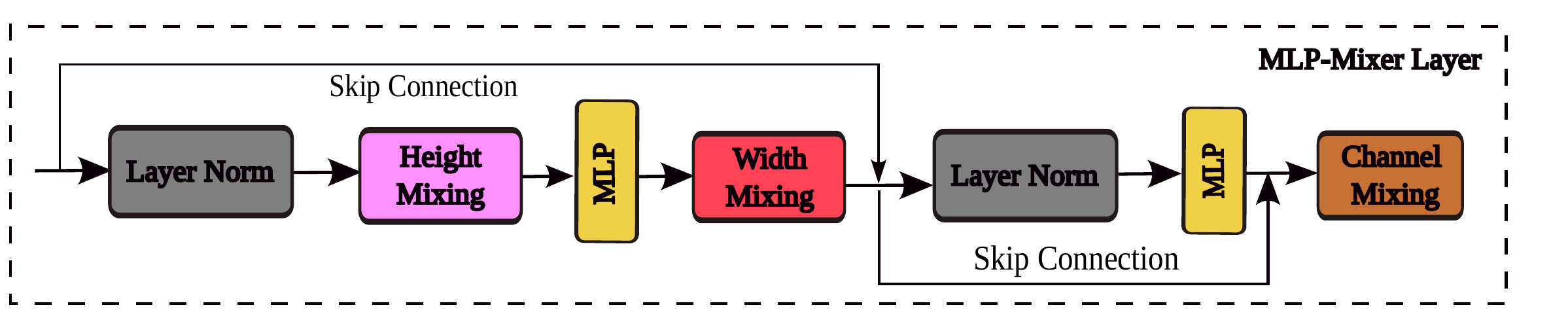}
\caption{Graphical representation of MLP-Mixer layer.}
\label{fig:prop2}
\end{figure*}

As illustrated in Fig.~\ref{fig:prop}, the SGU-MLP, is developed for image classification using a small number of training data. For efficient application of the multi-scale representation in the classification task, we incorporated a computationally light and straightforward depth-wise CNN-based architecture. As presented in Fig.~\ref{fig:prop2}, the MLP-Mixer layer of the developed model includes two different types of layers: (i) MLPs utilized across image patches for extraction of spatial information and (ii) MLPs utilized individually to extract per-location features from image inputs. In addition, in each MLP block, the SGU is utilized to enable the developed algorithm to effectively learn intricate spatial relationships among the tokens of the input data.
\subsection{Depth-wise Convolution Block (DWC):} The DWC architecture is light and straightforward, and is based on CNNs. With so many variables and the limited available training data, a higher probability of overfitting exists during the training process. Hence, to address overfitting and capture multi-scale feature information, we incorporated three depth-wise convolutions in parallel. These convolutions consist of filters with a size of 20 and kernel ($k$) sizes of $1\times{1}$, $3\times{3}$, and $5\times{5}$, respectively. Feature maps $X$ with a size of $9\times{9}\times{d}$ are the input for the DWC block that produces output $D_Z$, where $d$ is the number of bands. 

\begin{equation}
D_Z = \sum_{j=1,3,5}\rm{DWConv2D_{(k\times{k})}(\textit{X})}  
\end{equation}

The output maps of the three depth-wise CNNs are added and fed to the MLP-Mixer blocks.

\subsection{Spatial gating unit (SGU):} The SGU is designed to extract complex spatial interaction across tokens. Unlike, the current ViT models, the SGU does not necessitate the use of positional embedding. In other words, the positional embedding information is obtained through the use of spatial depth-wise convolutions \cite{Liub35} similar to inverted bottlenecks employed in MobileNetV2 \cite{Sandler_2018_CVPR}. Considering the dense layer of $D$ in the MLP block, as illustrated in Fig.~\ref{fig:prop}, the SGU uses a linear projection layer that benefits from a contraction operation across the spatial dimension of the cross-tokens interaction as defined by:
\begin{equation}
f_{W,b}(D) = W D + b
\end{equation}
\noindent where $W\in R^{n\times{n}}$ defines a matrix that has a size equal to the input sequence length, while $n$ and $b$ present the sequence length and biases of the tokens. It should be highlighted that the spatial projection matrix of $W$ is not dependent on the input data, contradicting the self-attention models where $W(D)$ is created dynamically from the $D$. The SGU can be formulated as:
\begin{equation}
S(D) = D \cdot f_{W,b}(D)
\end{equation}
\noindent where element-wise multiplication is represented by $(\cdot)$. The SGU equation can be improved by dividing $D$ into $D1$ and $D2$ along the channel dimension. Thus, the SGU can be formulated as:
\begin{equation}
S(D) = D1 \cdot f_{W,b}(D2)
\end{equation}

The output map of the DWC block is flattened and fed to the MLP-Mixer layer. Considering a dense layer of size $256\times{256}$, The $D1$ and $D2$ both have sizes of $256\times{128}$. The $f_{W,b}(D2)$ has a size of $256\times{128}$, where the $S(D)$ has a size of $256\times{128}$.

\subsection{Multi-layer Perceptron Mixer Block (MLP-Mixer):} In current advanced deep vision architectures, layers combine features in one or more of the following ways:  (1) at a given spatial location, (2) among various spatial locations, or (3) both simultaneously, with kernel of $k\times{k}$ convolutions (for $k$ $>$ 1) and pooling operations (2), incorporated in CNNs.  Convolutions with kernel size $1\times{1}$ perform the operation (1), whereas convolutions with larger kernels accomplish both operations (1) and (2). Self-attention layers in ViTs and other attention-based structures include operations (1) and (2), while models based on the MLPs only perform (1). The objective of the MLP-Mixer architecture is to distinguish between cross-location (height and width mixing) operations and per-location (channel-mixing) operations, as presented in Fig.~\ref{fig:prop2} \cite{Tolsti2021}. A series of non-overlapping patches of images $E$ from the output feature of the DWC block $D_Z$ are the input to the MLP-Mixer that is projected to a given hidden dimension of $C$, resulting in two-dimensional data of $\mathbf{M}\in\mathcal{R}^{E\times{d}}$, where $d$ illustrates the image input band number. Given the input image of size $H\times{W}$, and patches of $F\times{F}$, the number of patches would be $E=\frac{H\times{W}}{F^2}$, where all resulting patches of images are projected into the same projection matrix. The MLP-Mixer consists of several identical size layers, where each layer has two MLP blocks. The first token-mixing (i.e., height and width mixing) MLP block is applied on the columns of $M$, while the second MLP block (i.e., channel mixing) is utilized on the rows of the $M$. Two fully connected layers are in each MLP block, and a non-linearity function is applied independently to each row of the input image tensors. As such, each MLP-Mixer can be formulated as:
\begin{equation}
U_{\iota,i} = M_{\iota,i} + W_2 \xi (W_1 LN(M)_{\iota,i})), i= 1, ..., B
\end{equation}
\begin{equation}
Y_{j,\iota} = U_{j,\iota} + W_4 \xi (W_3 LN(U)_{\iota,i})), j= 1, \ldots, E
\end{equation}

Notably, the MLP-Mixer has a linear computation complexity, which distinguishes it from vision transformers with quadratic computation complexity and, consequently, exhibits a high level of computational efficiency.\\

\subsection{Spatial Gating Unit Multi-layer Perceptron (SGU-MLP):} 

Let us consider three data modalities, $X_1$, $X_2$, and $X_3$. From these datasets, image patches with the size of $9\times{9}$ are extracted and then concatenated. As seen in Fig.~\ref{fig:prop}, the concatenated layer is fed to the DWC layer. After being fed into the DWC block, the input images of size $9\times{9}\times{B}$ result in equal feature maps of size $9\times{9}\times{B}$, where $B$ represents the number of bands. The resulting feature map is then flattened and passed on to the MLP-Mixer blocks. The MLP-Mixer includes four blocks with patch sizes of 4, token dimension of 256, and channel dimension of 256. As discussed, in each MLP block, before the activation function (i.e., GELU), the SGU is employed to extract complex spatial interactions between the tokens. Finally, the last layer of the MLP-Mixer is a dense layer with a softmax activation function. The size of the last layer is equal to the number of existing classes in each study area.

\section{Experimental Results}
\label{sec:exp}

\subsection{Experimental Data}

\textbf{Houston dataset:} 
This dataset was captured over the University of Houston campus and the neighboring urban area. It consists of a coregistered hyperspectral and multispectral dataset containing 144 and 8 bands, respectively, with $349\times{1905}$ pixels. More information can be found at \cite{6776408}.

\textbf{Berlin dataset:} This dataset has a spatial resolution of $797\times{220}$ pixels and contains 244 spectral bands with wavelengths ranging from 0.4 $\mu$m to 2.5 $\mu$m over Berlin. The Sentinel-1 dual-Pol (VV-VH) single-look complex (SLC) product represents the SAR data. The processed SAR data have a satial resolution of $1723\times{476}$ pixels and a 13.89 m GSD. The HS image is nearest neighbor interpolated, as for the Houston dataset, to provide the same image size as the SAR data \cite{Okujeni2016}.

\textbf{Augsburg dataset:} This scene over the city of Augsburg, Germany includes three distinct datasets: a spaceborne HS image, a dual-Pol PolSAR image and a DSM image. The PolSAR data were obtained from the Sentinel-1 platform, and the HS and DSM data were obtained by DAS-EOC and DLR. All image spatial resolutions were downscaled to a single 30 m GSD. The scene describes four features from the dual-Pol (VV-VH) SAR image, 180 spectral bands from 0.4 $\mu$m to 2.5 $\mu$m for the HS image and one DSM image of  $332\times{485}$ pixels \cite{HONG202168}.

\subsection{Classification Results}

\begin{figure*}[!htbp]
\centering
\begin{subfigure}{.3\columnwidth}
\centering
\includegraphics[clip=true, trim = 200 200 300 100, width=0.7\linewidth]{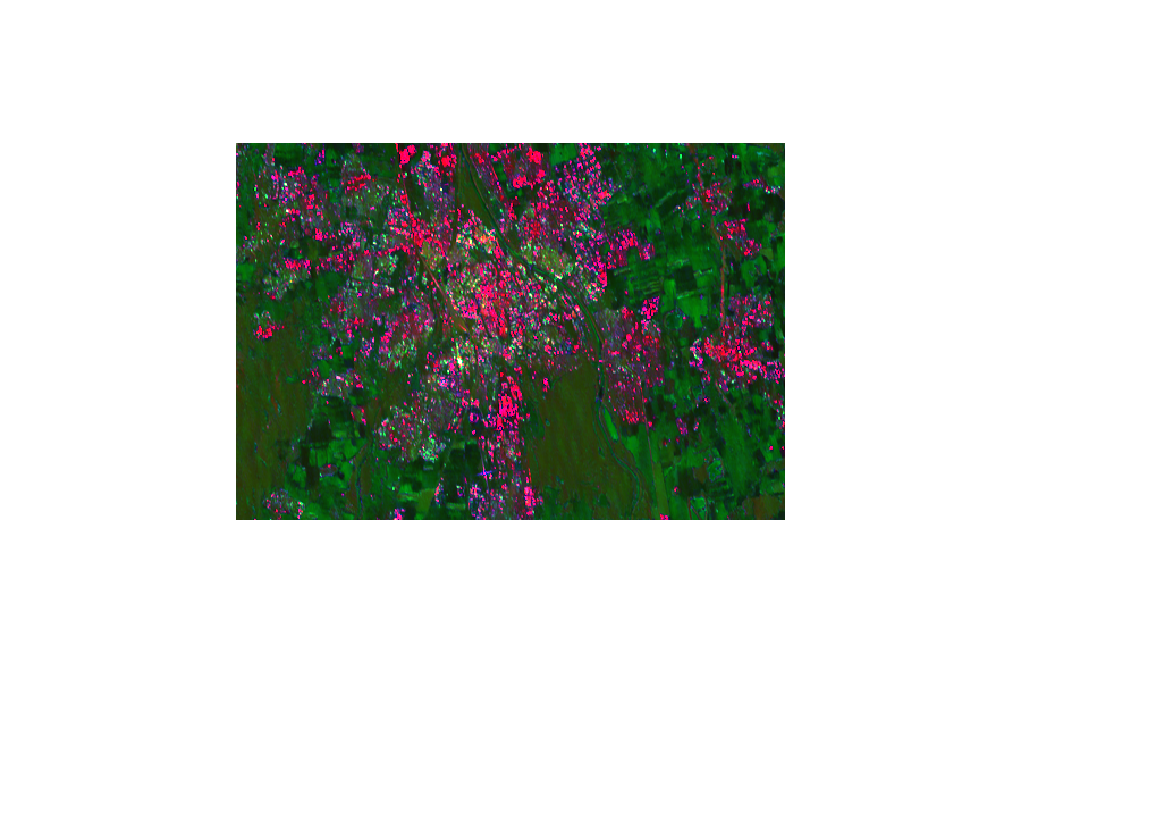} 
\caption{}
\end{subfigure}%
\begin{subfigure}{.3\columnwidth}
\centering
\includegraphics[clip=true, trim = 200 200 300 100, width=0.7\linewidth]{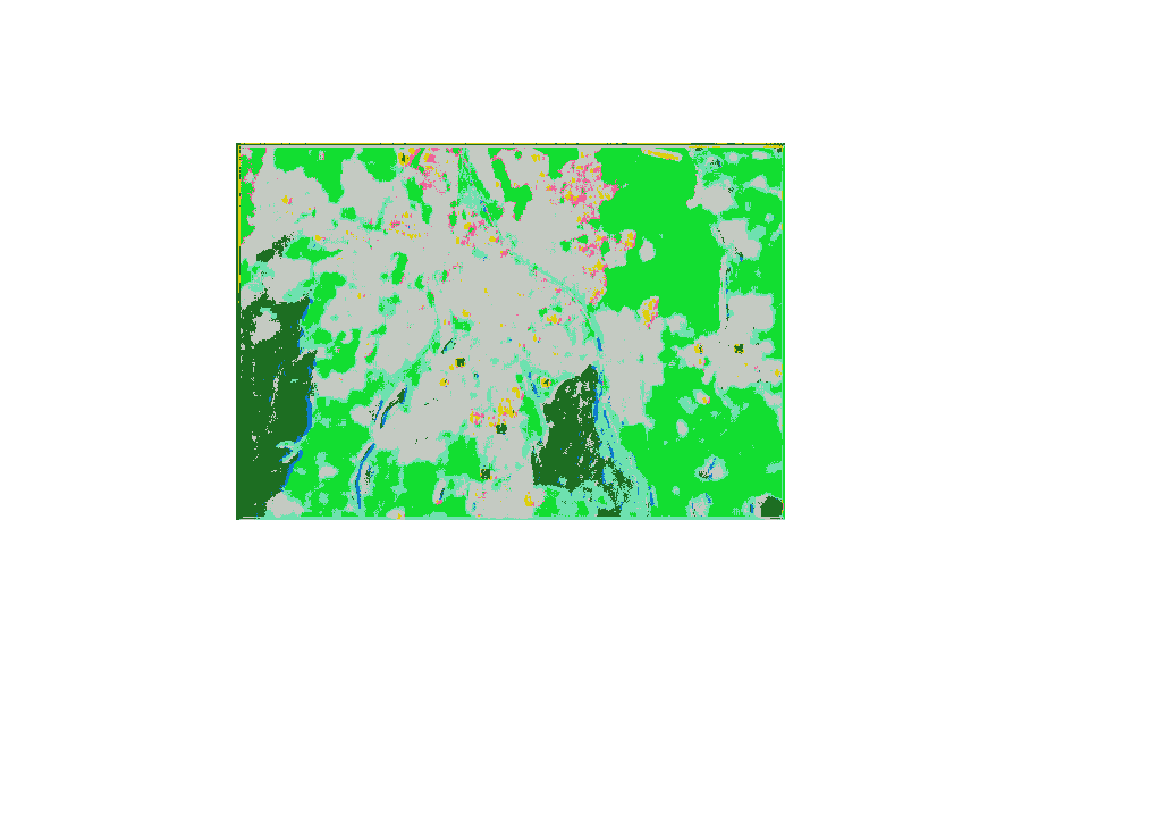} 
\caption{}
\end{subfigure}%
\begin{subfigure}{.3\columnwidth}
\centering
\includegraphics[clip=true, trim = 200 200 300 100, width=0.7\linewidth]{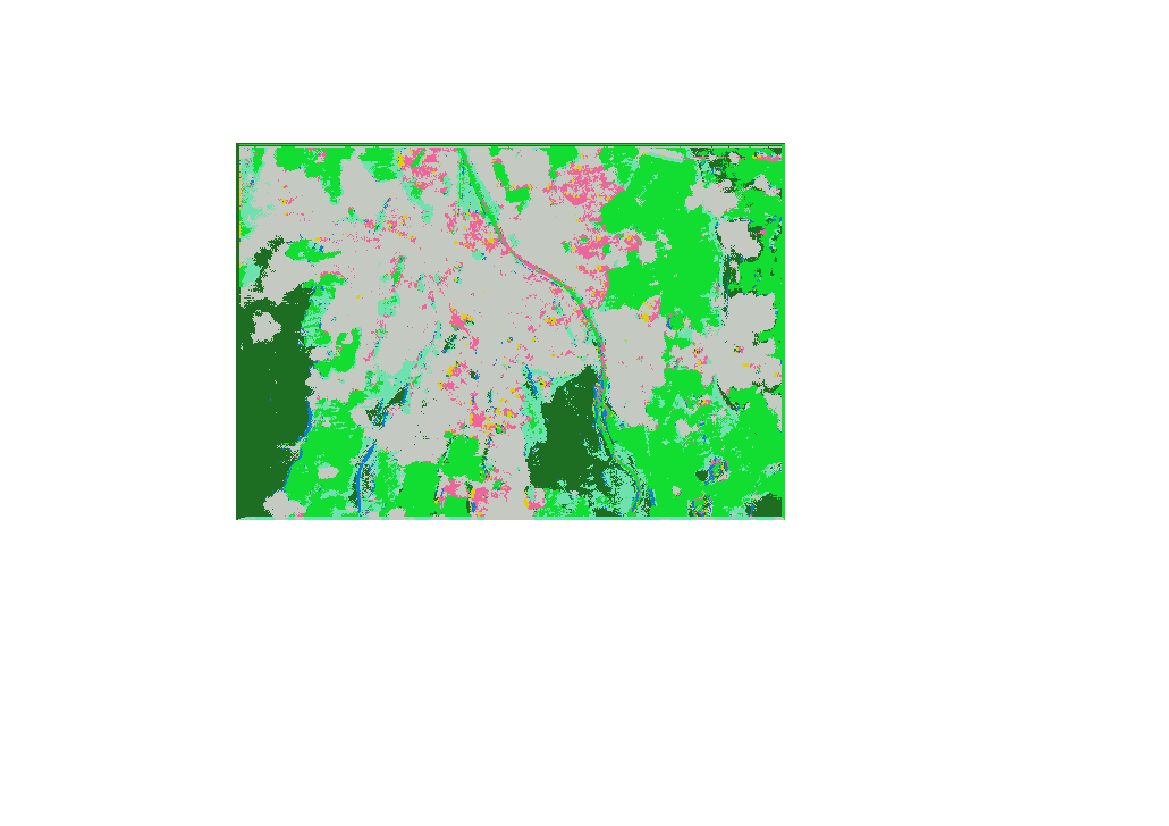} 
\caption{}
\end{subfigure}%
\begin{subfigure}{.3\columnwidth}
\centering
\includegraphics[clip=true, trim = 200 200 300 100, width=0.7\linewidth]{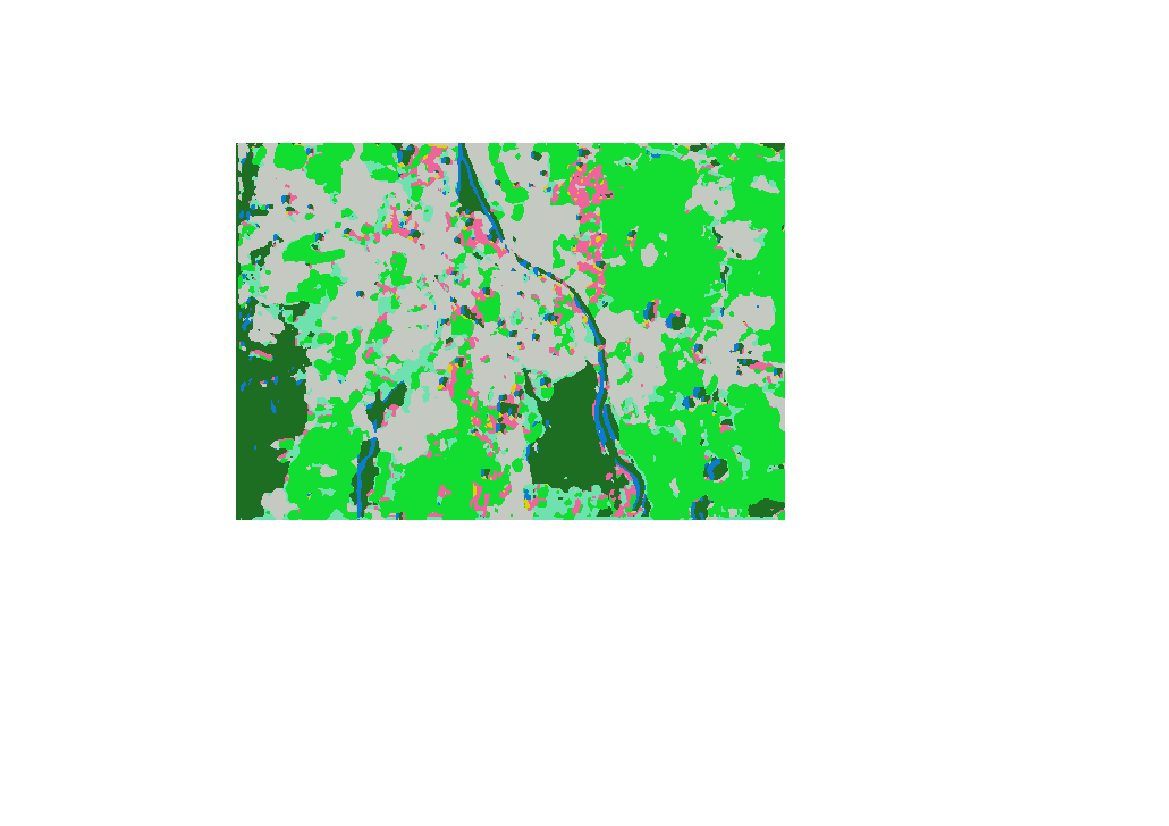} 
\caption{}
\end{subfigure}%
\begin{subfigure}{.3\columnwidth}
\centering
\includegraphics[clip=true, trim = 200 200 300 100, width=0.7\linewidth]{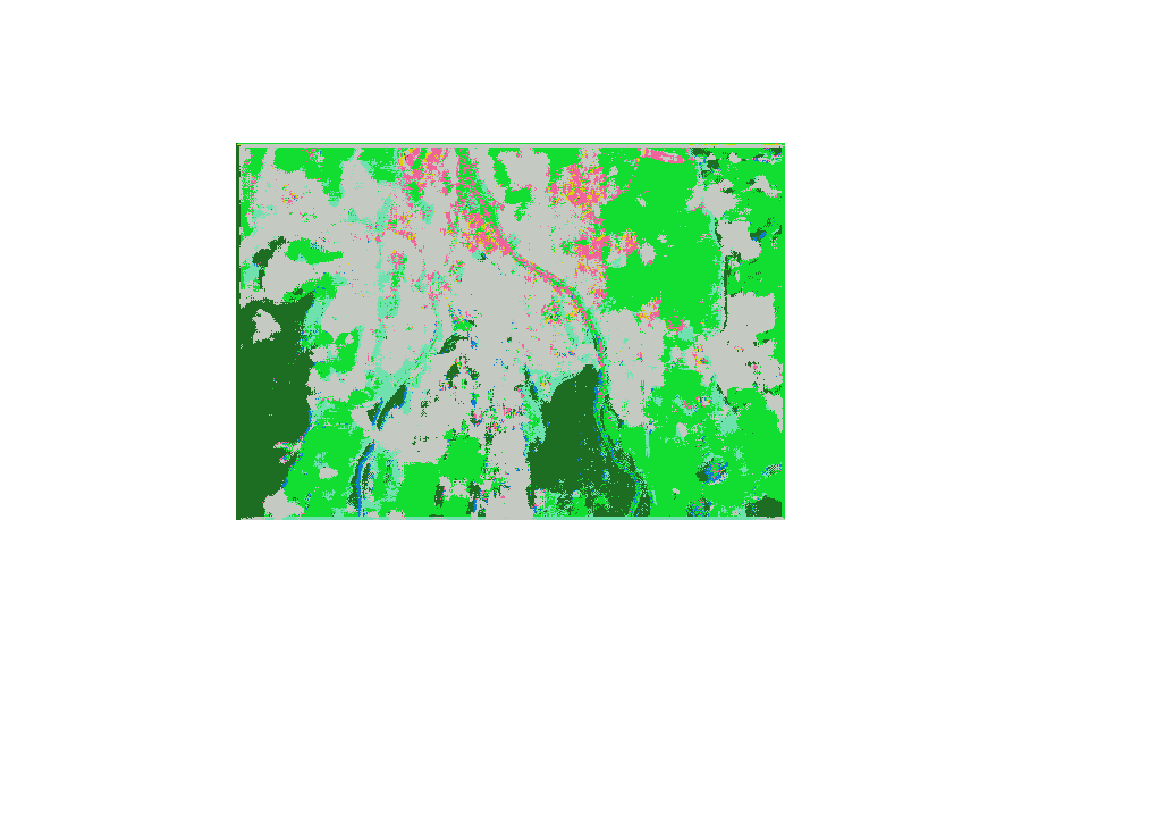} 
\caption{}
\end{subfigure}%
\begin{subfigure}{.3\columnwidth}
\centering
\includegraphics[clip=true, trim = 200 200 300 100, width=0.7\linewidth]{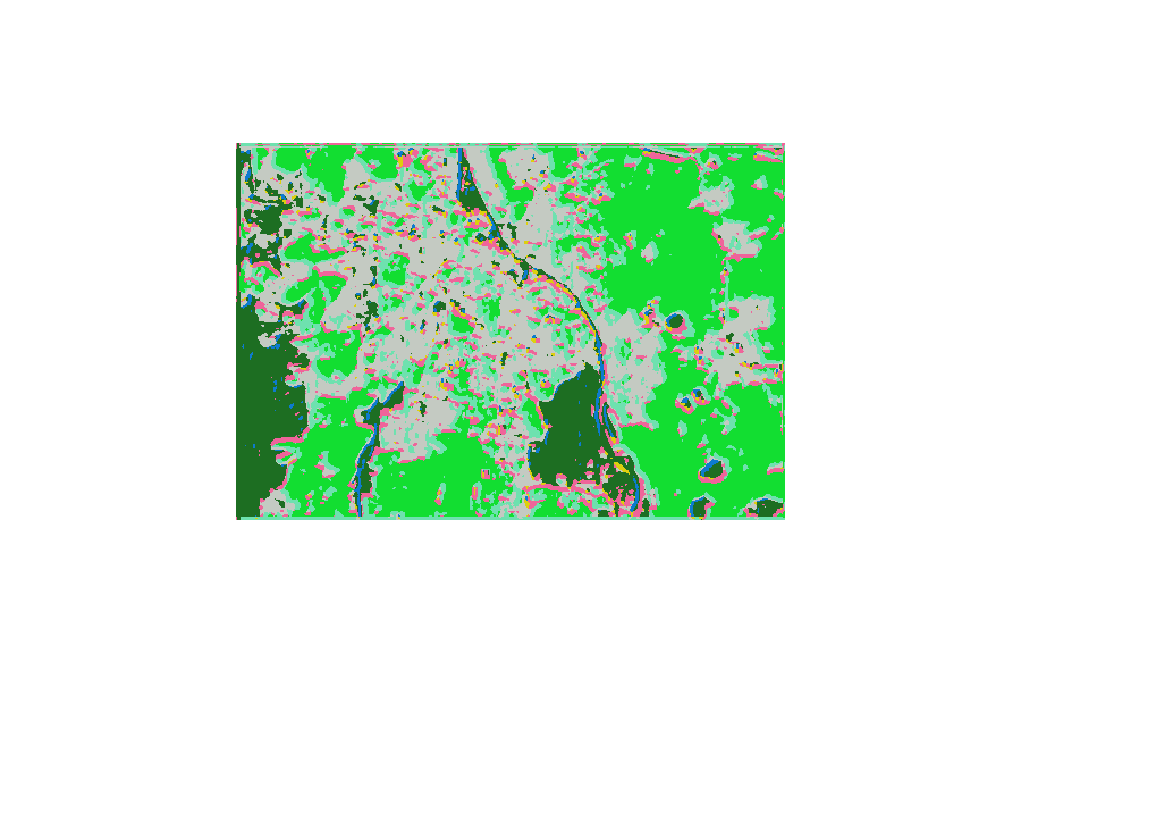} 
\caption{}
\end{subfigure}%
\begin{subfigure}{.3\columnwidth}
\centering
\includegraphics[clip=true, trim = 200 200 300 100, width=0.7\linewidth]{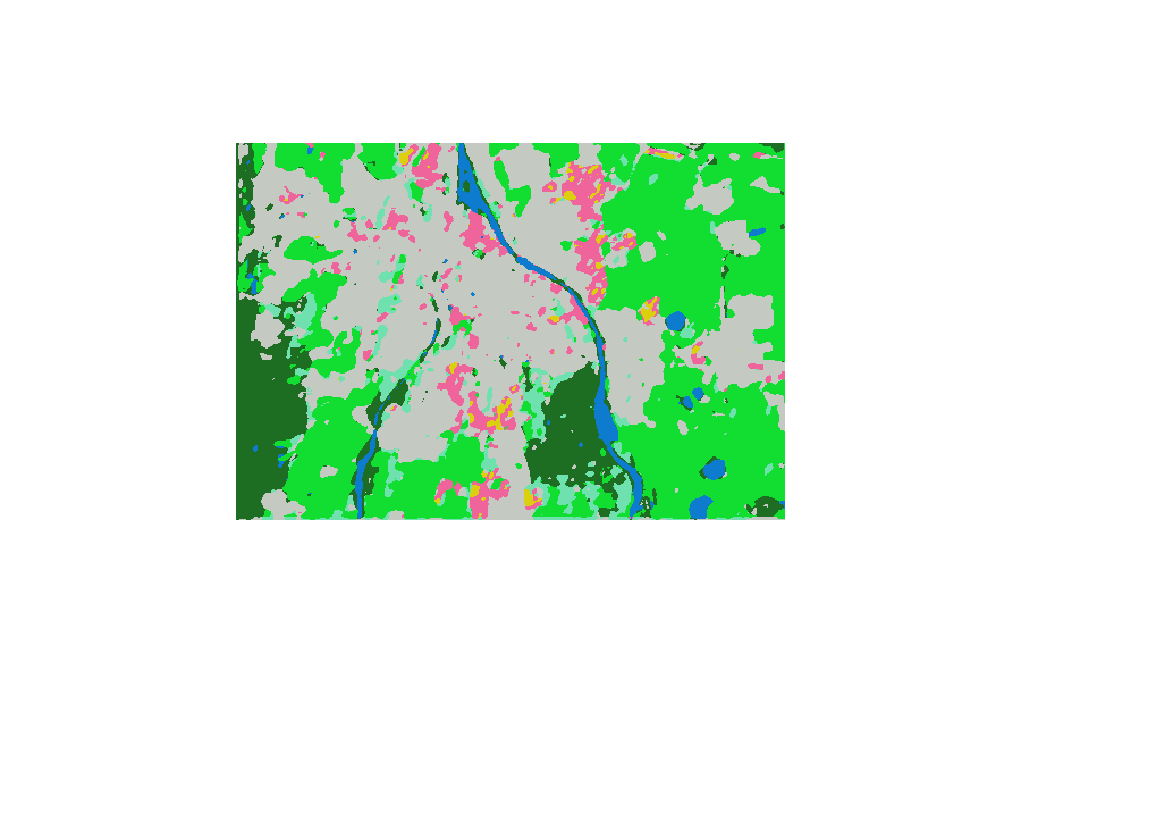} 
\caption{}
\end{subfigure}%
\\
\fcolorbox{forest!}{forest!}{\null} Forest
\fcolorbox{residential!}{residential!}{\null} Residential
\fcolorbox{industrial!}{industrial!}{\null} Industrial
\fcolorbox{lowplants!}{lowplants!}{\null} Low plants
\fcolorbox{allotment!}{allotment!}{\null} Allotment
\fcolorbox{water!}{water!}{\null} Water
\caption{Classification Maps over the Augsburg dataset using a) Study image, b) CoAtNet, c) EfficientFormer, d) HybridSN, e) iFormer, f) ResNet, and g) the SGU-MLP.}
\label{fig:map_Aug}
\end{figure*}

The classification capability of the developed SGU-MLP was evaluated against several CNN-based and cutting-edge CNN-ViT algorithms, including HybridSN \cite{roy2019hybridsn}, ResNet \cite{He_2016_CVPR}, iFormer \cite{Zhou_2021}, EfficientFormer \cite{li2022} and CoAtNet \cite{dai_coatnet_2021}. In the Augsburg dataset, as seen in Table~\ref{tab:Augsburg}, the developed SGU-MLP algorithm demonstrated superior classification performance with an average accuracy of 66.79\% compared to ResNet (43.57\%), CoAtNet(49.9\%), Efficientformer (52.81\%), iFormer (52.96\%) and HybridSN (55.76\%). The developed SGU-MLP classifier significantly increased the classification accuracy of the CNN-ViT-based algorithms of iFormer,	Efficientformer, and CoAtNet by about 21\%, 21\%, and 25\% in terms of average accuracy, as illustrated in Table~\ref{tab:Augsburg} and Fig.~\ref{fig:map_Aug}.

\begin{figure*}[!htbp]
\centering
\begin{subfigure}{.35\columnwidth}
\centering
\includegraphics[clip=true, trim = 0 0 0 0, angle=90,  width=0.95\linewidth]{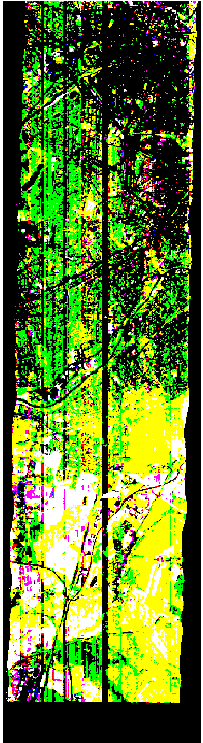} 
\caption{}
\end{subfigure}%
\begin{subfigure}{.35\columnwidth}
\centering
\includegraphics[clip=true, trim = 0 0 0 0, angle=90, width=0.95\linewidth]{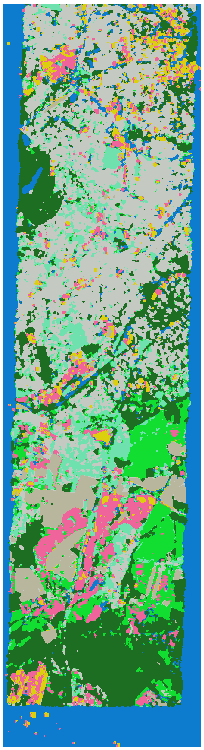} 
\caption{}
\end{subfigure}%
\begin{subfigure}{.35\columnwidth}
\centering
\includegraphics[clip=true, trim = 0 0 0 0, angle=90, width=0.95\linewidth]{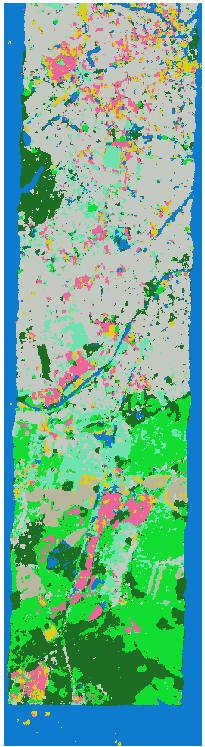} 
\caption{}
\end{subfigure}%
\begin{subfigure}{.35\columnwidth}
\centering
\includegraphics[clip=true, trim = 0 0 0 0, angle=90,  width=0.95\linewidth]{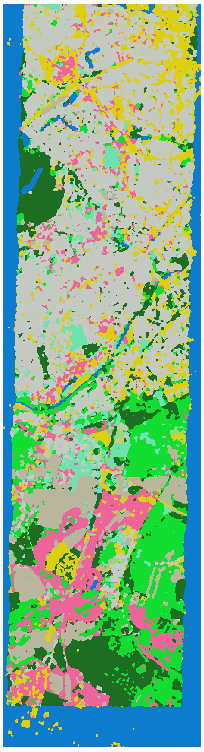} 
\caption{}
\end{subfigure}%
\begin{subfigure}{.35\columnwidth}
\centering
\includegraphics[clip=true, trim =0 0 0 0, angle=90, width=0.95\linewidth]{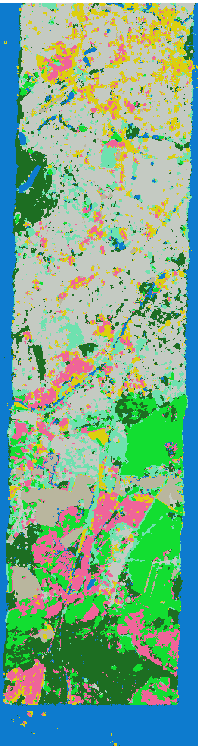} 
\caption{}
\end{subfigure}%
\\
\begin{subfigure}{.35\columnwidth}
\centering
\includegraphics[clip=true, trim = 0 0 0 0, angle=90, width=0.95\linewidth]{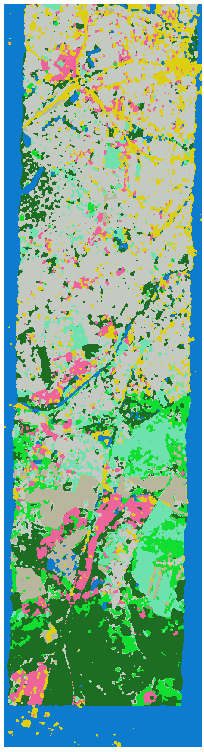} 
\caption{}
\end{subfigure}%
\begin{subfigure}{.35\columnwidth}
\centering
\includegraphics[clip=true, trim = 0 0 0 0, angle=90,  width=0.95\linewidth]{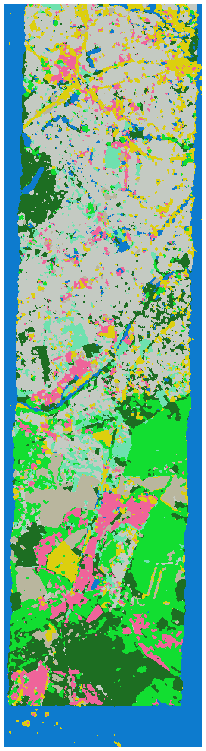} 
\caption{}
\end{subfigure}%
\\
\fcolorbox{forest!}{forest!}{\null} Forest
\fcolorbox{residential!}{residential!}{\null} Residential
\fcolorbox{industrial!}{industrial!}{\null} Industrial
\fcolorbox{soil!}{soil!}{\null} Soil
\fcolorbox{lowplants!}{lowplants!}{\null} Low plants
\fcolorbox{allotment!}{allotment!}{\null} Allotment
\fcolorbox{water!}{water!}{\null} Water
\caption{Classification Maps over the Berlin dataset using a) Study image, b) CoAtNet, c) EfficientFormer, d) HybridSN, e) iFormer, f) ResNet, and g) the SGU-MLP.}
\label{fig:map_Berlin}
\end{figure*}

In the Berlin study area, the SGU-MLP classifier with an average accuracy of 66.26\% considerably increased the classification accuracy of the other CNN-ViT algorithms iFormer, CoAtNet and Efficientformer by approximately 5\%, 9\% and 9\%, respectively, as shown in Table~\ref{tab:Berlin} and Fig.~\ref{fig:map_Berlin}. The MLP-SGU achieved F-1 scores of 0.27, 0.41, 0.46, 0.48, 0.67, 0.72, 0.73 and 0.82 for the recognition of commercial areas, industrial areas, allotment, water, soil, low plants, forest and residential areas, respectively.

\begin{figure*}[!htbp]
\centering
\begin{subfigure}{0.35\columnwidth}
\centering
\includegraphics[clip=true, trim = 0 0 0 0, width=0.95\linewidth]{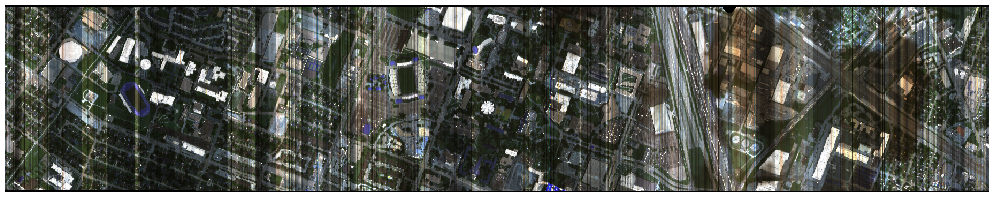} 
\caption{}
\end{subfigure}%
\begin{subfigure}{0.35\columnwidth}
\centering
\includegraphics[clip=true, trim = 0 0 0 0, width=0.95\linewidth]{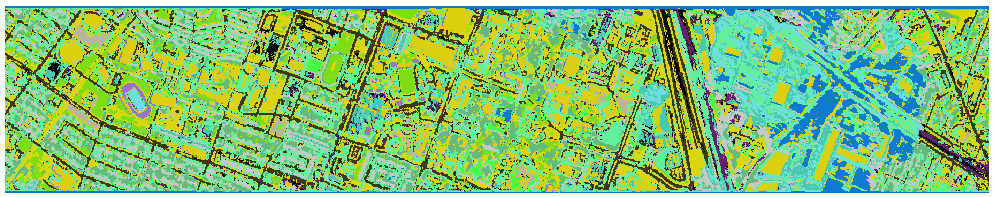} 
\caption{}
\end{subfigure}%
\begin{subfigure}{0.35\columnwidth}
\centering
\includegraphics[clip=true, trim = 0 0 0 0, width=0.95\linewidth]{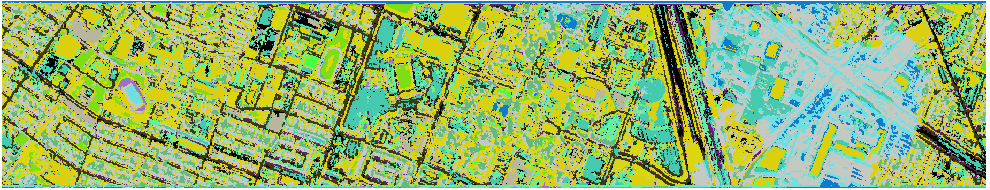} 
\caption{}
\end{subfigure}%
\begin{subfigure}{0.35\columnwidth}
\centering
\includegraphics[clip=true, trim = 0 0 0 0, width=0.95\linewidth]{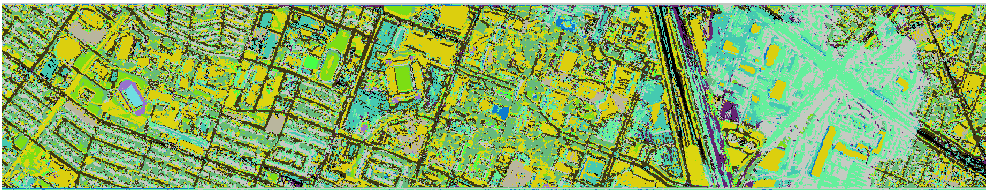} 
\caption{}
\end{subfigure}%
\begin{subfigure}{0.35\columnwidth}
\centering
\includegraphics[clip=true, trim =0 0 0 0, width=0.95\linewidth]{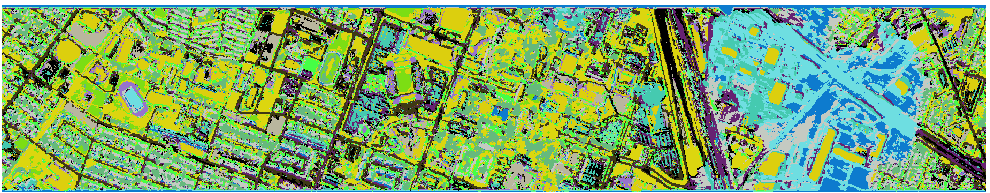} 
\caption{}
\end{subfigure}%
\\
\begin{subfigure}{0.35\columnwidth}
\centering
\includegraphics[clip=true, trim = 0 0 0 0, width=0.95\linewidth]{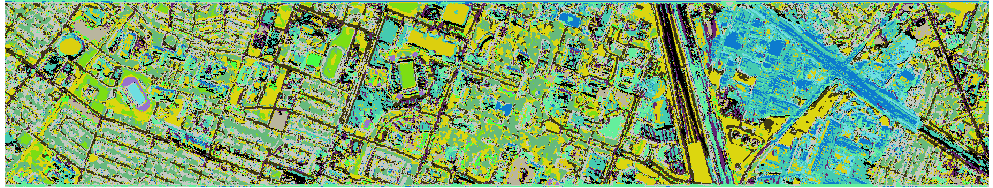} 
\caption{}
\end{subfigure}%
\begin{subfigure}{0.35\columnwidth}
\centering
\includegraphics[clip=true, trim = 0 0 0 0, width=0.95\linewidth]{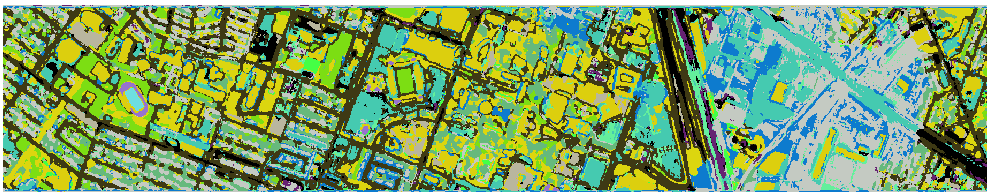} 
\caption{}
\end{subfigure}%
\\
\fcolorbox{residential!}{residential!}{\null} Residential
\fcolorbox{allotment!}{allotment!}{\null} Allotment
\fcolorbox{water!}{water!}{\null} Water
\fcolorbox{commercial!}{commercial!}{\null} Commercial
\fcolorbox{healthygrass!}{healthygrass!}{\null} Healthy grass
\fcolorbox{stressedgrass!}{stressedgrass!}{\null} Stressed grass
\fcolorbox{syntheticgrass!}{syntheticgrass!}{\null} Synthetic grass
\fcolorbox{tree!}{tree!}{\null} Tree
\fcolorbox{road!}{road!}{\null} Road
\fcolorbox{highway!}{highway!}{\null} Highway
\fcolorbox{railway!}{railway!}{\null} Railway
\fcolorbox{parkingL1!}{parkingL1!}{\null} Parking Lot1
\fcolorbox{parkingL2!}{parkingL2!}{\null} Parking Lot2
\fcolorbox{tenniscourt!}{tenniscourt!}{\null} Tennis court
\fcolorbox{runningtrack!}{runningtrack!}{\null} Running track
\caption{Classification Maps over the Houston dataset using a) Study image, b) CoAtNet, c) EfficientFormer, d) HybridSN, e) iFormer, f) ResNet, and g) the SGU-MLP.}
\label{fig:map_Houston}
\end{figure*}

As shown in Table~\ref{tab:Houston} and Fig.~\ref{fig:map_Houston}, with a kappa index of 86.91\%, the SGU-MLP algorithm noticeably surpassed the classification performance of the ResNet (65.49\%), iFormer (68.71\%), Efficientformer (69.25\%), CoAtNet (70.56\%), and HybridSN (73.59\%), respectively, in the Houston pilot site. The developed SGU-MLP classification model outperformed the other CNN and CNN-ViT-based algorithms of the HybridSN, CoAtNet, Efficientformer, iFormer, and ResNet  by about 15\%, 19\%, 20\%, 21\%, and 25\%, respectively, in terms of average accuracy, as demonstrated in Table~\ref{tab:Houston}.

\begin{table}[!ht]
\centering
\caption{Classification results of Augsburg dataset in terms of F-1 score where $\kappa$ = Kappa index, OA = Overall Accuracy, AA = Average Accuracy, respectively.}
\resizebox{0.85\linewidth}{!}{
\begin{tabular}{|c|c|c|c|c|c|c|} \hline
Class &	HybridSN & ResNet & iFormer	& Efficientformer & CoAtNet & SGU-MLP
\\
 \hline
Forest &	0.88 &	0.83	& 0.91 &	0.88 &	0.85 &	\textbf{0.94}\\ 
Residential &	0.89 &	0.83 &	0.89 &	0.9 &	0.87 &	\textbf{0.95} \\ 
Industrial  & 0.43 &	0.15 &	0.35 &	0.4 &	0.22 &	\textbf{0.63}\\ 
Low Plants &	0.87 &	0.88 &	0.88 &	0.88 &	\textbf{0.98} &	0.96\\
Allotment &	 0.13 &	0.1 &	0.13 &	0.11 &	0.09 &	\textbf{0.36}\\  
Commercial &	 0.04 &	0.05 &	0.1	& 0.11 &	0.16 &	\textbf{0.28}\\  
Water &	 0.35 &	0.19 &	0.21 &	0.25 &	0.19 &	\textbf{0.52}\\  
\hline \hline
OA$\times$100 & 82.28 &	79.07	& 82.82 &	82.72	& 81.32 &	\textbf{91.82}\\
AA$\times$100 & 55.76	& 43.57	& 52.96 &	52.81 &	49.9 &	\textbf{66.79}\\
$\kappa\times$100 & 74.85 &	69.34 &	75.37 &	75.24 &	73.12 &	\textbf{88.22}\\
\hline
\end{tabular}}
\label{tab:Augsburg}
\end{table}

\begin{table}[!ht]
\centering
\caption{Classification results of Berlin dataset in terms of F-1 score where $\kappa$ = Kappa index, OA = Overall Accuracy, AA = Average Accuracy, respectively.}
\resizebox{0.85\linewidth}{!}{
\begin{tabular}{|c|c|c|c|c|c|c|} \hline
Class &	HybridSN & ResNet & iFormer	& Efficientformer & CoAtNet & SGUMLP
\\
 \hline
Forest & 0.71 &	0.64 &	0.69 &	\textbf{0.73} &	0.65 &	\textbf{0.73}	\\ 
Residential & 0.8 &	0.81 &	\textbf{0.82} &	0.81 &	0.76 &	\textbf{0.82}	 \\ 
Industrial  & \textbf{0.49} &	0.39 &	0.35 &	0.32 &	0.32 &	0.41 \\ 
Low Plants & 0.59 &	0.35 &	0.72 &	0.7 &	0.59 &	\textbf{0.72}	\\
Soil &	 0.65 &	0.72 &	0.7 &	0.67 &	\textbf{0.75} &	0.67\\  
Allotment &	0.44 &	0.28 &	0.34 &	0.29 &	0.3 &	\textbf{0.46} \\  
Commercial &	\textbf{0.45} &	0.25 &	0.29 &	0.24 &	0.29 &	0.27 \\  
Water &	 \textbf{0.65} & 0.53 &	0.49 &	0.38 &	0.28 &	0.48\\  
\hline \hline
OA$\times$100 & 66.81 &	63.7 &	68.6 &	68.17 &	63.14 &	\textbf{70.79}\\
AA$\times$100 & 62.67 &	58.23 &	62.84 &	60.05 &	60.53 &	\textbf{66.26}\\
$\kappa\times$100 & 55.84 &	47.61 &	55.28 &	54.32 &	49.21 &	\textbf{58.06}\\
\hline
\end{tabular}}
\label{tab:Berlin}
\end{table}

\begin{table}[!ht]
\centering
\caption{Classification results of Houston dataset in terms of F-1 score where $\kappa$ = Kappa index, OA = Overall Accuracy, AA = Average Accuracy, respectively.}
\resizebox{0.85\linewidth}{!}{
\begin{tabular}{|c|c|c|c|c|c|c|} \hline
Class &	HybridSN & ResNet & iFormer	& Efficientformer & CoAtNet & SGUMLP
\\
 \hline
Healthy Grass & 	0.85 &	0.88 &	0.86 &	0.89 &	\textbf{0.9} &	0.89\\ 
Stressed Grass & 	 0.84 &	\textbf{0.9} &	0.87 &	0.87 &	0.88 &	0.89
\\ 
Synthetic Grass  &  0.84 &	0.78 &	0.5 &	0.58 &	0.72 &	\textbf{0.97}
\\ 
Tree & 	0.87 &	0.89 &	0.92 &	0.91 &	0.93 &	\textbf{0.94}
\\
Soil &	 0.96 &	0.94 &	0.93 &	0.95 &	0.85 &	\textbf{0.99}
\\  
Water &	 \textbf{0.73} &	0.71 &	0.29 &	0.39 &	0.25 &	0.35
\\  
Residential &	 0.69 &	0.72 &	0.68 &	0.6 &	0.79 &	\textbf{0.81}
\\  
Commercial &	 0.69 &	0.39 &	0.68 &	0.56 &	0.6 &	\textbf{0.83}
\\  
Road & 	0.7	& 0.57 &	0.75 &	0.77 &	0.82 &	\textbf{0.85}
\\ 
Highway & 	 0.58 &	0.52 &	0.45 &	0.54 &	0.54 &	\textbf{0.89}
\\ 
Railway  &  0.7 &	0.54 &	0.67 &	0.57 &	0.67 &	\textbf{0.82}
\\ 
Parking Lot1 & 	0.74 &	0.42 & 0.48	& 0.71 &	0.55 &	\textbf{0.96}
\\
Parking Lot2 &	\textbf{0.94} &	0.61 &	0.72 &	0.78 &	0.58 &	0.9
 \\  
Tennis Court &	 0.84 &	0.77 &	0.74 &	0.73 &	0.56 &	\textbf{0.99}
\\  
Running Track &	 0.64 &	0.82 &	0.83 &	0.61 &	0.92 &	\textbf{0.95}
\\  

\hline \hline
OA$\times$100 & 75.62 &	68.16 &	71.03 &	71.66 &	72.67
 &	\textbf{87.91}\\
AA$\times$100 & 76.44 &	71.42 &	72.86 &	70.69 &	75.62
 &	\textbf{89.33}\\
$\kappa\times$100 & 73.59 &	65.49 &	68.71 &	69.25 &	70.56
 &	\textbf{86.91}\\
\hline
\end{tabular}}
\label{tab:Houston}
\end{table}

\subsection{Ablation study}

An ablation study was performed to better understand the contribution and significance of different parts of the developed SGU-MLP classification algorithm. As seen in Table~\ref{tab:Ablation1}, the inclusion of the DWC block and SGU block increased the classification accuracy of the MLP-Mixer model by approximately 4\% and 1\%, respectively, in terms of average accuracy for the Augsburg dataset. The highest classification accuracy was achieved by the inclusion of both the DWC and SGU blocks with an average accuracy of 66.79\%, increasing the classification accuracy of the MLP-Mixer algorithm by about 8\%.

In the Berlin dataset, as illustrated in Table~\ref{tab:Ablation2}, the inclusion of the SGU block and DWC block increased the classification accuracy of the MLP-Mixer algorithm by about 3\% and 4\%, respectively, in terms of Kappa index. By incorporating both the DWC and SGU blocks, the highest classification was attained with a Kappa index of 58.06\%. This increased the accuracy of the MLP-Mixer classifier by approximately 10\%.

As demonstrated in Table~\ref{tab:Ablation3}, the inclusion of the DWC block and SGU block increased the accuracy of the MLP-Mixer algorithm by approximately 1\% and 3\%, respectively, in terms of average accuracy for the Houston dataset. By the inclusion of both the DWC and SGU blocks, the MLP-Mixer's classification accuracy was increased by approximately 9\% to 89.33\%.

\begin{table}[!ht]
\centering
\caption{Classification results of Augsburg dataset in terms of F-1 score where $\kappa$ = Kappa index, OA = Overall Accuracy, AA = Average Accuracy, respectively.}
\resizebox{0.7\linewidth}{!}{
\begin{tabular}{|c|c|c|c|c|} \hline
Class &	MLP & SGU + MLP & DWC + MLP	 & SGUMLP
\\
 \hline
Forest &	0.88 &	0.92 &	\textbf{0.94} &	\textbf{0.94}\\ 
Residential &	0.93 &	\textbf{0.95} &	\textbf{0.95} &	\textbf{0.95} \\ 
Industrial  & 0.57 &	0.57 &	0.54 &	\textbf{0.63}\\ 
Low Plants &	0.96 &	\textbf{0.98} &	\textbf{0.98} &	\textbf{0.98}\\
Allotment &	  0.17 &	0.27 &	\textbf{0.41} &	0.36\\  
Commercial &	0.19 &	0.26 &	0.25 &	\textbf{0.28}\\  
Water &	 \textbf{0.56} &	\textbf{0.56} &	0.53 &	0.52\\  
\hline \hline
OA$\times$100 &  88.21 &	90.86 &	 91.39 &	\textbf{91.82}\\
AA$\times$100 &   61.41 &	65.83 &	64.42 &	\textbf{66.79}\\
$\kappa\times$100 &  82.95 &	86.83 &	87.57 &	\textbf{88.22}\\
\hline
\end{tabular}}
\label{tab:Ablation1}
\end{table}

\begin{table}[!ht]
\centering
\caption{Classification results of Augsburg dataset in terms of F-1 score where $\kappa$ = Kappa index, OA = Overall Accuracy, AA = Average Accuracy, respectively.}
\resizebox{0.7\linewidth}{!}{
\begin{tabular}{|c|c|c|c|c|} \hline
Class &	MLP & SGU + MLP & DWC + MLP	 & SGUMLP
\\
 \hline
Forest &	0.71 &	\textbf{0.74} &	0.70 &	0.73
\\ 
Residential &	 0.78 &	0.80 &	0.81 &	\textbf{0.82}
\\ 
Industrial  & 0.34 &	0.38 &	0.40 &	\textbf{0.41}
\\ 
Low Plants &	0.61 &	0.70 &	0.66 &	\textbf{0.72}
\\
Soil &	 0.66 &	\textbf{0.70} &	\textbf{0.70} &	0.67
 \\  
Allotment &	 0.40 &	0.43 &	\textbf{0.46} &	\textbf{0.46}
\\  
Commercial &	 \textbf{0.31} &	0.28 &	0.25 &	0.27
\\  
Water &	 0.43 &	0.47 &	\textbf{0.53} &	0.48
\\  
\hline \hline
OA$\times$100 &  65.47 &	68.96 &	69.11
 &	\textbf{70.79}\\
AA$\times$100 &     64.11 &	66.2 &	65.07
 &	\textbf{66.26}\\
$\kappa\times$100 &  52 &	56.1 &	55.88
 &	\textbf{58.06}\\
\hline
\end{tabular}}
\label{tab:Ablation2}
\end{table}

\begin{table}[!ht]
\centering
\caption{Classification results of Houston dataset in terms of F-1 score where $\kappa$ = Kappa index, OA = Overall Accuracy, AA = Average Accuracy, respectively.}
\resizebox{0.7\linewidth}{!}{
\begin{tabular}{|c|c|c|c|c|} \hline
Class &	MLP & SGU + MLP & DWC + MLP	 & SGUMLP
\\
 \hline
Healthy Grass & 0.89 &	\textbf{0.90} &	\textbf{0.90} &	\textbf{0.90}	\\ 
Stressed Grass & 0.89 &	0.89 &	\textbf{0.90}	 & 0.89	 \\ 
Synthetic Grass  & 0.69 &	0.96 &	\textbf{0.97} &	\textbf{0.97} \\ 
Tree &  0.90 &	0.93 &	0.92 &	\textbf{0.94}	\\
Soil &	0.95 &	\textbf{1} &	\textbf{1} &	0.99\\  
Water &	0.66 &	0.21 &	\textbf{0.81} &	0.35 \\  
Residential &	0.65 &	0.80 &	0.77 &	\textbf{0.81}\\  
Commercial & 0.70 &	0.82 &	0.81 &	\textbf{0.83}	 \\  
Road & 	0.79 &	0.87 &	0.81 &	\textbf{0.85}\\ 
Highway & 	 0.54 &	0.74 &	0.68 &	\textbf{0.89}\\ 
Railway  &  0.71 &	\textbf{0.84} &	0.82 &	0.82\\ 
Parking Lot1 & 	0.76 &	\textbf{0.96} &	0.95 &	\textbf{0.96}\\
Parking Lot2 &	0.89 &	0.89 &	0.87 &	\textbf{0.90}\\  
Tennis Court &	0.93 &	\textbf{1} &	\textbf{1} &	0.99 \\  
Running Track &	0.90 &	\textbf{0.96} &	0.95 &	0.95 \\  

\hline \hline
OA$\times$100 & 78.04 &	85.03 &	86.44  &	\textbf{87.91}\\
AA$\times$100 &  81.26 &	87.03 &	88.12 &	\textbf{89.33}\\
$\kappa\times$100 &  76.20 &	83.85 &	85.28 &	\textbf{86.91}\\
\hline
\end{tabular}}
\label{tab:Ablation3}
\end{table}

\section{Conclusion}
\label{sec:con}

Convolutional Neural Networks (CNNs) are commonly utilized frameworks for hierarchical feature extraction. At the same time, due to the use of a self-attention system, vision transformers (ViTs) can achieve better modeling of global contextual information than CNNs. However, to realize their image classification capability, ViTs require large training datasets. To overcome this limitation, we developed the SGU-MLP algorithm based on advanced MLP models and a spatial gating unit for land use land cover mapping which demonstrated superior classification accuracy compared to several CNN and CNN-ViT-based models. For the Houston experiment, for example, with a Kappa index of 86.91\%, the SGU-MLP algorithm significantly outperformed the classification accuracy of the ResNet (65.49\%), iFormer (68.71\%), Efficientformer (69.25\%), CoAtNet (70.56\%) and HybridSN (73.59\%) algorithms. For the Augsburg dataset, the SGU-MLP algorithm, with an average accuracy of 66.26\%, again demonstrated increased classification accuracy compared to ResNet (43.57\%), CoAtNet(49.9\%), Efficientformer (52.81\%), iFormer (52.96\%) and HybridSN (55.76\%).

\bibliographystyle{IEEEtran}
\bibliography{reference}

\begin{thebibliography}{10}
\providecommand{\url}[1]{#1}
\csname url@samestyle\endcsname
\providecommand{\newblock}{\relax}
\providecommand{\bibinfo}[2]{#2}
\providecommand{\BIBentrySTDinterwordspacing}{\spaceskip=0pt\relax}
\providecommand{\BIBentryALTinterwordstretchfactor}{4}
\providecommand{\BIBentryALTinterwordspacing}{\spaceskip=\fontdimen2\font plus
\BIBentryALTinterwordstretchfactor\fontdimen3\font minus
  \fontdimen4\font\relax}
\providecommand{\BIBforeignlanguage}[2]{{%
\expandafter\ifx\csname l@#1\endcsname\relax
\typeout{** WARNING: IEEEtran.bst: No hyphenation pattern has been}%
\typeout{** loaded for the language `#1'. Using the pattern for}%
\typeout{** the default language instead.}%
\else
\language=\csname l@#1\endcsname
\fi
#2}}
\providecommand{\BIBdecl}{\relax}
\BIBdecl

\bibitem{Yang680}
\BIBentryALTinterwordspacing
J.~Yang, A.~Guo, Y.~Li, Y.~Zhang, and X.~Li, ``Simulation of landscape spatial
  layout evolution in rural-urban fringe areas: a case study of ganjingzi
  district,'' \emph{GIScience \& Remote Sensing}, vol.~56, no.~3, pp. 388--405,
  2019. [Online]. Available:
  \url{https://doi.org/10.1080/15481603.2018.1533680}
\BIBentrySTDinterwordspacing

\bibitem{HAASE201292}
\BIBentryALTinterwordspacing
D.~Haase, A.~Haase, N.~Kabisch, S.~Kabisch, and D.~Rink, ``Actors and factors
  in land-use simulation: The challenge of urban shrinkage,''
  \emph{Environmental Modelling \& Software}, vol.~35, pp. 92--103, 2012.
  [Online]. Available:
  \url{https://www.sciencedirect.com/science/article/pii/S1364815212000606}
\BIBentrySTDinterwordspacing

\bibitem{FAN20249}
\BIBentryALTinterwordspacing
Y.~Fan, X.~Ding, J.~Wu, J.~Ge, and Y.~Li, ``High spatial-resolution
  classification of urban surfaces using a deep learning method,''
  \emph{Building and Environment}, vol. 200, p. 107949, 2021. [Online].
  Available:
  \url{https://www.sciencedirect.com/science/article/pii/S036013232100353X}
\BIBentrySTDinterwordspacing

\bibitem{Ghamisi2156}
P.~Ghamisi, B.~Rasti, N.~Yokoya, Q.~Wang, B.~Hofle, L.~Bruzzone, F.~Bovolo,
  M.~Chi, K.~Anders, R.~Gloaguen, P.~M. Atkinson, and J.~A. Benediktsson,
  ``Multisource and multitemporal data fusion in remote sensing: A
  comprehensive review of the state of the art,'' \emph{IEEE Geoscience and
  Remote Sensing Magazine}, vol.~7, no.~1, pp. 6--39, 2019.

\bibitem{ahmad2021hyperspectral}
A.~Muhammad, S.~Sidrah, R.~Swalpa~Kumar, H.~Danfeng, W.~Xin, Y.~Jing, M.~K.
  Adil, M.~Manuel, D.~Salvatore, and C.~Jocelyn, ``Hyperspectral image
  classification--traditional to deep models: A survey for future prospects,''
  \emph{IEEE Journal of Selected Topics in Applied Earth Observations and
  Remote Sensing}, vol.~15, pp. 968--999, 2022.

\bibitem{roy2022multimodal}
S.~K. Roy, A.~Deria, D.~Hong, B.~Rasti, A.~Plaza, and J.~Chanussot,
  ``Multimodal fusion transformer for remote sensing image classification,''
  \emph{arXiv preprint arXiv:2203.16952}, 2022.

\bibitem{Tolsti2021}
I.~O. Tolstikhin, N.~Houlsby, A.~Kolesnikov, L.~Beyer, X.~Zhai, T.~Unterthiner,
  J.~Yung, A.~Steiner, D.~Keysers, J.~Uszkoreit, M.~Lucic, and A.~Dosovitskiy,
  ``Mlp-mixer: An all-mlp architecture for vision,'' in \emph{Advances in
  Neural Information Processing Systems}, M.~Ranzato, A.~Beygelzimer,
  Y.~Dauphin, P.~Liang, and J.~W. Vaughan, Eds., vol.~34.\hskip 1em plus 0.5em
  minus 0.4em\relax Curran Associates, Inc., 2021, pp. 24\,261--24\,272.

\bibitem{Liub35}
\BIBentryALTinterwordspacing
H.~Liu, Z.~Dai, D.~So, and Q.~V. Le, ``Pay attention to mlps,'' in
  \emph{Advances in Neural Information Processing Systems}, M.~Ranzato,
  A.~Beygelzimer, Y.~Dauphin, P.~Liang, and J.~W. Vaughan, Eds., vol.~34.\hskip
  1em plus 0.5em minus 0.4em\relax Curran Associates, Inc., 2021, pp.
  9204--9215. [Online]. Available:
  \url{https://proceedings.neurips.cc/paper_files/paper/2021/file/4cc05b35c2f937c5bd9e7d41d3686fff-Paper.pdf}
\BIBentrySTDinterwordspacing

\bibitem{Sandler_2018_CVPR}
M.~Sandler, A.~Howard, M.~Zhu, A.~Zhmoginov, and L.-C. Chen, ``Mobilenetv2:
  Inverted residuals and linear bottlenecks,'' in \emph{Proceedings of the IEEE
  Conference on Computer Vision and Pattern Recognition (CVPR)}, June 2018.

\bibitem{6776408}
C.~Debes, A.~Merentitis, R.~Heremans, J.~Hahn, N.~Frangiadakis, T.~van
  Kasteren, W.~Liao, R.~Bellens, A.~Pižurica, S.~Gautama, W.~Philips,
  S.~Prasad, Q.~Du, and F.~Pacifici, ``Hyperspectral and lidar data fusion:
  Outcome of the 2013 grss data fusion contest,'' \emph{IEEE Journal of
  Selected Topics in Applied Earth Observations and Remote Sensing}, vol.~7,
  no.~6, pp. 2405--2418, 2014.

\bibitem{Okujeni2016}
A.~Okujeni, S.~van~der Linden, and P.~Hostert, ``Berlin-urban-gradient dataset
  2009 - an enmap preparatory flight campaign,'' 2016.

\bibitem{HONG202168}
\BIBentryALTinterwordspacing
D.~Hong, J.~Hu, J.~Yao, J.~Chanussot, and X.~X. Zhu, ``Multimodal remote
  sensing benchmark datasets for land cover classification with a shared and
  specific feature learning model,'' \emph{ISPRS Journal of Photogrammetry and
  Remote Sensing}, vol. 178, pp. 68--80, 2021. [Online]. Available:
  \url{https://www.sciencedirect.com/science/article/pii/S0924271621001362}
\BIBentrySTDinterwordspacing

\bibitem{roy2019hybridsn}
S.~K. Roy, G.~Krishna, S.~R. Dubey, and B.~B. Chaudhuri, ``Hybridsn: Exploring
  3-d--2-d cnn feature hierarchy for hyperspectral image classification,''
  \emph{IEEE Geoscience and Remote Sensing Letters}, vol.~17, no.~2, pp.
  277--281, 2019.

\bibitem{He_2016_CVPR}
K.~He, X.~Zhang, S.~Ren, and J.~Sun, ``Deep residual learning for image
  recognition,'' in \emph{Proceedings of the IEEE Conference on Computer Vision
  and Pattern Recognition (CVPR)}, June 2016.

\bibitem{Zhou_2021}
\BIBentryALTinterwordspacing
H.~Zhou, S.~Zhang, J.~Peng, S.~Zhang, J.~Li, H.~Xiong, and W.~Zhang,
  ``Informer: Beyond efficient transformer for long sequence time-series
  forecasting,'' \emph{Proceedings of the AAAI Conference on Artificial
  Intelligence}, vol.~35, no.~12, pp. 11\,106--11\,115, May 2021. [Online].
  Available: \url{https://ojs.aaai.org/index.php/AAAI/article/view/17325}
\BIBentrySTDinterwordspacing

\bibitem{li2022}
Y.~Li, G.~Yuan, Y.~Wen, J.~Hu, G.~Evangelidis, S.~Tulyakov, Y.~Wang, and
  J.~Ren, ``Efficientformer: Vision transformers at mobilenet speed,'' 2022.

\bibitem{dai_coatnet_2021}
Z.~Dai, H.~Liu, Q.~Le, and M.~Tan, ``{CoAtNet}: Marrying convolution and
  attention for all data sizes,'' in \emph{Advances in Neural Information
  Processing Systems 34}, 2021.

\end{thebibliography}

\end{document}